\documentclass[sigconf, screen,9pt,authorversion,nonacm]{acmart}

\usepackage{amsmath,amsfonts}
\usepackage{array}
\usepackage[caption=false,font=normalsize,labelfont=sf,textfont=sf]{subfig}

\usepackage{textcomp}
\usepackage{stfloats}
\usepackage{url}
\usepackage{verbatim}
\usepackage{amsmath}
\usepackage{graphicx}
\usepackage{siunitx}
\usepackage[noend]{algpseudocode}
\usepackage{setspace}
\usepackage{fancyhdr}
\usepackage{ifthen}
\usepackage{xcolor}
\usepackage{upgreek}
\usepackage{fontawesome5}
\usepackage{layouts}
\usepackage[capitalise]{cleveref}
\usepackage{booktabs}
\usepackage{multirow}
\usepackage{xspace}
\usepackage{changepage}
\usepackage{soul}
\usepackage{acro}
\usepackage{tikz}
\usepackage{pgfplots}
\usepackage{pgfplotstable}
\usepgfplotslibrary{fillbetween}
\usepgfplotslibrary{groupplots}
\usepgfplotslibrary{statistics}
\usepgfplotslibrary{colorbrewer}
\usepackage{comment}
\usepackage{makecell} 

\usepackage{algorithm}
\usepackage[noend]{algpseudocode}
\usepackage{bm}
\usepackage{enumitem}

\usepackage{diagbox}

\pgfplotsset{compat=1.18}
\usetikzlibrary{arrows,calc,patterns,intersections,positioning,shapes,shapes.symbols,shapes.geometric,arrows.meta,fit,backgrounds}

\DeclareAcronym{DVFS}{short=DVFS,long=dynamic voltage and frequency scaling}
\DeclareAcronym{IPS}{short=IPS,long=instructions per second}
\DeclareAcronym{RL}{short=RL,long=reinforcement learning}
\DeclareAcronym{RM}{short=RM,long=resource management}
\DeclareAcronym{LOO}{short=LOO,long=leave-one-out}
\DeclareAcronym{IPJ}{short=IPJ,long=instructions per Joule}
\DeclareAcronym{HMP}{short=HMP,long=heterogeneous multi-core processor}
\DeclareAcronym{LLC}{short=LLC,long=last-level cache,short-indefinite=an}
\DeclareAcronym{ML}{short=ML,long=machine learning,short-indefinite=an}
\DeclareAcronym{NN}{short=NN,long=neural network,short-indefinite=an}
\DeclareAcronym{DNN}{short=DNN,long=deep neural network,short-indefinite=an}
\DeclareAcronym{MCS}{short=MCS,long=mixed-criticality systems,short-indefinite=a}
\DeclareAcronym{MSE}{short=MSE,long=mean squared error}
\DeclareAcronym{MC}{short=MC,long=memory controller}
\DeclareAcronym{RMSE}{short=RMSE,long=root-mean-square error}
\DeclareAcronym{MAPE}{short=MAPE,long=mean-absolute-percentage error,short-indefinite=an}
\DeclareAcronym{ROI}{short=ROI,long=region of interest}
\DeclareAcronym{TDP}{short=TDP,long=Thermal Design Power}
\DeclareAcronym{TSP}{short=TSP,long=Thermal Safe Power}
\DeclareAcronym{DSE}{short=DSE,long=Design Space Exploration}
\DeclareAcronym{VF}{short=V/f,long=voltage/frequency}
\DeclareAcronym{ANN}{short=ANN,long=Artificial Neural Network}
\DeclareAcronym{PMC}{short=PMC,long=Performance Monitoring Counter}
\DeclareAcronym{GMM}{short=GMM,long=Gaussian Mixture Model}
\DeclareAcronym{aoi}{short=AoI, long= Application of Interest}
\DeclareAcronym{TCC}{short=TCC, long= Thermal Control Circuitry }
\DeclareAcronym{DTM}{short=DTM, long= Dynamic Thermal Management }
\DeclareAcronym{HBM}{short=HBM, long= High-Bandwidth Memory}
\DeclareAcronym{RDT}{short=HBM, long= Resource Director Technology}
\DeclareAcronym{CAT}{short=HBM, long= Cache Allocation Technology}
\DeclareAcronym{MBA}{short=HBM, long= Memory Bandwidth Allocation}
\DeclareAcronym{6G}{short=6G, long=6th Generation Mobile Networks}
\DeclareAcronym{URLLC}{short=URLLC, long=Ultra-Reliable Low-Latency Communication}
\DeclareAcronym{mMTC}{short=mMTC, long=Massive Machine-Type Communication}
\DeclareAcronym{NFV}{short=NFV, long=Network Function Virtualization}
\DeclareAcronym{MEC}{short=MEC, long=Multi-Access Edge Computing}
\DeclareAcronym{QoS}{short=QoS, long=Quality of Service}

\DeclareAcronym{FeMBB}{short=FeMBB, long=Further-enhanced Mobile Broadband}
\DeclareAcronym{ERLLC}{short=ERLLC, long=Extremely Reliable Low-Latency Communication}
\DeclareAcronym{umMTC}{short=umMTC, long=Ultra-Massive Machine-Type Communication}
\DeclareAcronym{LDHMC}{short=LDHMC, long=Long-Distance and High-Mobility Communication}
\DeclareAcronym{ELPC}{short=ELPC, long=Extremely Low Power Communication}
\DeclareAcronym{AE}{short=AE, long=autoencoder}
\DeclareAcronym{KNN}{short=KNN, long=\(k\)-nearest neighbors}
\DeclareAcronym{OOD}{short=OOD, long=out-of-distribution}
\DeclareAcronym{LSTM}{short=LSTM, long=Long Short-Term Memory}
\DeclareAcronym{iid}{short=i.i.d, long=independent and identically distributed}
\DeclareAcronym{KDE}{short=KDE, long=kernel density estimates}

\pgfdeclarelayer{background}
\pgfdeclarelayer{foreground}
\pgfsetlayers{background,main,foreground}

\pgfplotsset{compat=1.13,
	/pgfplots/ybar legend/.style={
		/pgfplots/legend image code/.code={%
			\draw[##1,/tikz/.cd,yshift=-0.25em]
			(0cm,0cm) rectangle (3pt,0.6em);},
	},
	scale only axis,  
	xlabel near ticks,
	ylabel near ticks,
	xlabel style={font=\footnotesize,align=center},
	ylabel style={font=\footnotesize,align=center},
	xlabel style={yshift=1.5mm},
	ylabel style={yshift=-1.5mm},
	x tick label style={font=\footnotesize},
	y tick label style={font=\footnotesize},
	legend style={
		font={\footnotesize},
		fill=none,
	},
	legend cell align={left},
	max space between ticks=20pt  
}
\tikzset{
	container/.style={inner sep=0pt, outer sep=0pt},
	database/.style={
		draw,
		cylinder,
		shape border rotate=90,
		aspect=0.1,
	},
}

\makeatletter
\@ifpackageloaded{algpseudocode}{
	\algrenewcommand\algorithmicrequire{\textbf{Input:}}
	\algrenewcommand\algorithmicensure{\textbf{Output:}}
	\algrenewcommand\algorithmiccomment[1]{{\small \hfill\(\triangleright\) #1}}

	\algnewcommand{\algorithmicbreak}{\textbf{break}}
	\algnewcommand{\Break}{\State \algorithmicbreak}
	\algnewcommand{\algorithmiccontinue}{\textbf{continue}}
	\algnewcommand{\Continue}{\State \algorithmiccontinue}
}{}
\makeatother

\newcommand{\ourtech}{\emph{LEX}\xspace}
\newcommand{\ourtechplain}{\emph{Lightweight Runtime EXplainability}\xspace}

\newcommand*\circled[1]{\tikz[baseline=(char.base)]{\node[shape=circle,fill,inner sep=0.5pt] (char) {\textcolor{white}{#1}};}}

\makeatletter

\begin{document}

\title{XAI-Driven Confidence Estimation for OOD-Resilient Runtime Prediction}

\cfoot{This is the authors' preprint version of the work.} 

\author{Mahmoud Sanad}
\email{mahmoud.sanad@kit.edu}
\affiliation{%
  \institution{Karlsruhe Institute of Technology}
  \city{Karlsruhe}
  \country{Germany}
}

\author{Mohammed Bakr Sikal}
\email{bakr.sikal@kit.edu}
\affiliation{%
  \institution{Karlsruhe Institute of Technology}
  \city{Karlsruhe}
  \country{Germany}
}

\author{Osama Abboud}
\email{osama.abboud@huawei.com}
\affiliation{%
  \institution{Huawei Heisenberg Research Center}
  \city{Munich}
  \country{Germany}
}

\author{Heba Khdr}
\email{heba.khdr@kit.edu}
\affiliation{%
  \institution{Karlsruhe Institute of Technology}
  \city{Karlsruhe}
  \country{Germany}
}

\author{Jörg Henkel}
\email{henkel@kit.edu}
\affiliation{%
  \institution{Karlsruhe Institute of Technology}
  \city{Karlsruhe}
  \country{Germany}
}

\begin{abstract}
Machine learning-based resource management (RM) models degrade under out-of-distribution (OOD) inputs at runtime, yet existing approaches in the literature rely on a single global error margin that is overly conservative and fails to distinguish reliable predictions from catastrophic failures.
We propose \ourtechplain (\ourtech), the first post-hoc XAI framework for input-specific runtime confidence estimation in ML-based RM.
\ourtech clusters the embeddings of the training input space using a Gaussian Mixture Model and associates per-cluster prediction error bounds to input samples, enabling fine-grained runtime confidence signals without retraining.
Evaluated on a state-of-the-art LSTM-based IPS forecasting model on Intel Alder Lake, across 34 benchmark applications, \ourtech achieves \textasciitilde94\% containment on unseen workloads and produces \textasciitilde17\% tighter per-input confidence bounds than global baselines.
With its negligible runtime overhead, \ourtech advances the field by providing explainable and reliable input-specific confidence signals of ML predictions, thereby making ML-based resource management resilient to OOD at runtime.
\end{abstract}

\maketitle

\section{Introduction}
\label{sec:introduction}
\begin{figure}[t]
  \centering
  \includegraphics[width=\linewidth]{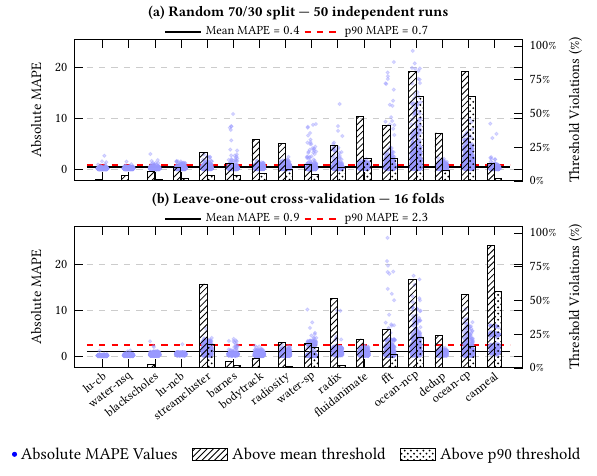}

  \caption{%
    Performance (in MAPE) of model inference on phase-level input data of different applications from the PARSEC/SPLASH-2 benchmark suites, with (a) 50 random 70/30 splits and (b) leave-one-out cross-validation.
    A significant percentage of data points exceed global thresholds, often exceeding 50\%, highlighting that global thresholds fail to provide a reliable confidence signal.
    }
  \label{fig:motivational}
\end{figure}

To cope with the growing complexity of modern multi-core processors and the rising demands for higher performance and efficiency, \ac{ML} has emerged as the dominant paradigm for system-level \ac{RM}.
In fact, a large body of \ac{ML} techniques has been proposed over the past decade, including supervised models that can predict workload behavior and guide \ac{DVFS}~\cite{Erika2022realted}, recurrent neural networks that forecast \ac{IPS} and \ac{IPJ} under resource contention~\cite{contribution_efficast}, and reinforcement and imitation learning models that derive resource management policies directly from hardware interactions~\cite{Narang_rw_cv_2023, costero2020rl}.
Collectively, these techniques have demonstrated significant improvements over heuristic baselines \cite{mlcad_martin_2022}, establishing \ac{ML}-based resource management as a promising research direction.

However, while very promising, \ac{ML} models face a major \textit{generalization} challenge when deployed in real-world settings.
Mainly, the shift between training and runtime data distributions exposes the model to \ac{OOD} samples, leading to performance degradation, i.e., catastrophic predictions, at runtime, potentially nullifying the intended benefits of \ac{ML}-based \ac{RM}.
Therefore, \iac{RM} technique shall be \textit{resilient} to such \ac{OOD} events at runtime, by accounting for model mispredictions, or even ignoring them in case of catastrophic predictions.
Yet, very few \ac{ML}-based techniques in the literature have demonstrated such awareness~\cite{Chen_rw_em_2023, Wachter_rw_em_2019, contribution_efficast, contribution_smartboost_dac_2021}.
These techniques have employed statistical metrics about the model performance or error, e.g., standard deviation from the \ac{MAPE}, 90-percentiles of \ac{MAPE}, etc., to adjust the model prediction at runtime by accounting for its error. 
However, as we show in the following motivational example, relying on such global statistical metrics hides the catastrophic performance a model can exhibit when faced with \ac{OOD} data samples at runtime.

\subsection{Motivational Example}
\label{sec:intro:motivational}

To quantify the severity of the \ac{OOD} problem, we evaluate the generalization of a state-of-the-art \ac{LSTM}-based \ac{IPS} forecasting model~\cite{contribution_efficast}.
To obtain training data, applications from the PARSEC and SPLASH-2 benchmark suites~\cite{parsec, splash2} are executed using the open-source ARDiS framework \cite{contribution_ardis} with their native input sets, on a Linux machine, having an Intel Alder Lake 24-core heterogeneous processor.

Each application is executed once on a performance ($P$) and once on an efficiency ($E$) core, at five \ac{VF} settings from 1.5 to 3.5 $\mathrm{GHz}$.
During each run, periodic traces are collected every 100 $\mathrm{ms}$, comprising the same CPU and cache/memory \acp{PMC} as in the original work, i.e., \textit{Instructions}, \textit{Branches}, \textit{LLC Loads}, \textit{Stores}, \textit{Branch Misses}, \textit{LLC Misses} and \textit{Store Misses}.
The raw periodic traces are then converted into structured multivariate time series for supervised sequence-to-one forecasting.
A sliding window of 5 time steps with one stride, i.e., the window moves one step at a time, generates overlapping subsequences, each paired with a target representing \ac{IPS} at the next step. 
All window sequences across all runs are combined into a single dataset containing \textasciitilde219K windows, with each window capturing the \textit{phase-level} characteristics of the application at a specific configuration, i.e., core type and \ac{VF} level.

Given the obtained dataset, the state-of-the-art \ac{LSTM}-based \ac{IPS} forecasting model is first trained then evaluated using the \ac{MAPE} score, following two strategies from the literature, and the results are shown on \cref{fig:motivational}.
For each experiment, the performance of the model on each data sample in the respective unseen test dataset is plotted as blue scattered dots, grouped by application.
In addition, the mean and the 90th-percentile (p90) \ac{MAPE} scores are plotted as black solid lines and red dashed lines, respectively.
 
\textbf{Experiment 1: Random split~(Fig.~\ref{fig:motivational}.a).}
In this experiment, the most common evaluation strategy in the literature is followed, which consists of splitting the dataset into random 70\%/30\% train/test splits.
With 50 different random splits, we train the model 50 times and compute the mean \ac{MAPE} scores across all 50 model variants.
As can be seen on Fig.~\ref{fig:motivational}.a, the model achieves a near-perfect mean \ac{MAPE} of $0.4$\% and a 90-th percentile \ac{MAPE} of $0.7$\% on the test dataset.
However, the spread of errors is severely wide for several applications, e.g., \textit{radix}, \textit{canneal}, \textit{fft}, and \textit{ocean-ncp}, with per-window \ac{MAPE} values ranging from near-zero to \textasciitilde24\%, depending on how the random split covers the applications' execution phases during training.
Importantly, this variance is \textit{not noise}, but aggregated over 50 trainings, which reflects a structural model sensitivity to different inputs, i.e., different execution phases of different applications.

This is expected; when a random split leaves a particular application's \textit{phase-level} characteristics under-represented in training, the model fails on those inputs at test time.
In extreme cases, certain execution phases are \textit{entirely absent} from training, leaving the model with no basis for reliable prediction, thereby aggravating the \ac{OOD} behavior.
To expose this behavior deterministically, we conduct a second \ac{LOO} experiment.

\textbf{Experiment 2: Leave-one-out cross-validation~(Fig.~\ref{fig:motivational}.b).}
In this experiment, we train 16 variants of the model in a \ac{LOO} protocol, each time holding out one application entirely from training, and using the same application for evaluating the model.
The results, shown in Fig.~\ref{fig:motivational}.b, confirm and amplify the pattern observed in Experiment~1; the global mean \ac{MAPE} rises to $0.9$\% and the 90th-percentile to $2.3\%$, with many high-variance samples that range from near-zero up to a catastrophic \textasciitilde26\% \ac{MAPE}.

\textbf{Critically}, across both experiments, the fraction of data samples that fall above the mean and p90 \ac{MAPE} scores exceeds 50\% for many applications.
This confirms our initial observation; relying on such global statistical metrics hides the catastrophic performance of the model when faced with \ac{OOD} data samples at runtime.
Therefore, global thresholds are \textit{unsufficient} to enable OOD-resilient ML-based resource management at runtime, thereby motivating the need for reliable confidence signals that are input specific. 

\subsection{Challenges and Contributions}
\label{sec:intro:challenges}

In both evaluation strategies in the motivational example, the model issues a scalar output regardless of whether it is operating on a well-covered or a catastrophically-uncovered input, i.e., \ac{OOD} samples.
Therefore, a \ac{RM} technique that employs such a model that fails silently at runtime would receive \textit{no signal} on which to base a decision to \textit{trust} or to \textit{override} the prediction.
Moreover, using a single global margin, i.e., mean or p90 of \ac{MAPE}, has proven insufficient as a model reliability signal.
In fact, there is a need for an \textit{input-dependent} reliability signal that reflects the \textit{per-input performance} of the model, delivering low error bounds for confident predictions and higher bounds for less certain ones.
If such a mechanism can capture the \textit{similarity} between different data samples, the performance of the model on some seen data samples can serve as an indicator of how the model would perform when faced with unseen ones at runtime, depending on the similarities between their features. 
However, designing such a mechanism is not trivial.

Mainly, the default features used in \ac{ML}-based \ac{RM} are \ac{PMC} readings, which are continuous values, i.e., non-discretized and non-categorical.
These readings attempt to capture high-level signatures of complex interactions between application characteristics, core microarchitecture, cache states, thermal conditions, \ac{VF} levels, etc.
Consequently, comparing individual features for similarity cannot capture such complex interactions.
For instance, two applications could exhibit the same \ac{IPS} behavior, but show significantly different cache access behavior.
Therefore, with such multi-dimensional continuous vectors, there is a need for \textit{learned representations}, capable of capturing the structural similarity of model behavior from the limited observations available during training.

Motivated by the aforementioned observations, this paper makes the following contributions:
\begin{itemize}
    \item[\circled{1}] We present the first systematic empirical analysis that evaluates the generalization behavior of \ac{ML}-based resource management techniques through a comprehensive set of experiments on real hardware, demonstrating that model failure is determined by phase-level characteristics of applications, that application-level \ac{LOO} evaluation systematically overestimates generalization, and that no single \ac{PMC} feature threshold suffices to detect \ac{OOD} inputs at runtime.
    \item[\circled{2}] We propose \ourtechplain~(\ourtech), a lightweight, yet robust, runtime confidence estimation framework for \ac{ML}-based resource management, that provides an explicit runtime reliability signal in the form of a quantitative, input-specific error margin, derived using \iac{GMM} fitted over a learned embedding space that captures structural similarity between execution phases. 
    \item[\circled{3}] We demonstrate through extensive experiments on a state-of-the-art resource management (\ac{RM}) forecasting model, that our adaptable input-specific confidence estimation technique successfully balances containment rate and error margins, across benchmarks and hardware configurations. Specifically, \ourtech delivers performance on par with the best global error margin while providing error bounds that are up to 10 times tighter, thereby showcasing the effectiveness of input-dependent confidence estimation in improving both the reliability and out-of-distribution (\ac{OOD}) resilience of ML-based resource management, all with negligible runtime overhead.
\end{itemize}

\section{Generalization Analysis}
\label{sec:analysis}

This section presents a comprehensive set of experiments conducted on real hardware, with the same experimental setup described in \cref{sec:intro:motivational}, considering the same state-of-the-art target model used in \cref{sec:intro:motivational}.
The model has an initial \ac{LSTM} layer with 64 units to capture temporal dependencies between features across the 5 epochs of the input window, followed by a second \ac{LSTM} layer with 32 units that produces a fixed-size latent vector encoding temporal patterns. 
This latent representation feeds a fully connected head of two dense layers (32 and 16 units with ReLU) and a single-neuron output for regression, i.e., \ac{IPS} value.
The model is trained with Adam (learning rate $1\times10^{-4}$), minimizing \ac{MSE}, using a batch size of 256 for up to 2000 epochs with early stopping (patience 10) to restore the best validation weights.

\subsection{Structural Determination of Model Failure}
\label{sec:analysis:failure}
\begin{figure*}
    
    \centering
    \begin{tikzpicture}[x=1cm,y=1cm,font=\footnotesize,scale=0.9]

      \node[inner sep=0pt] (img) {\includegraphics[width=\linewidth]{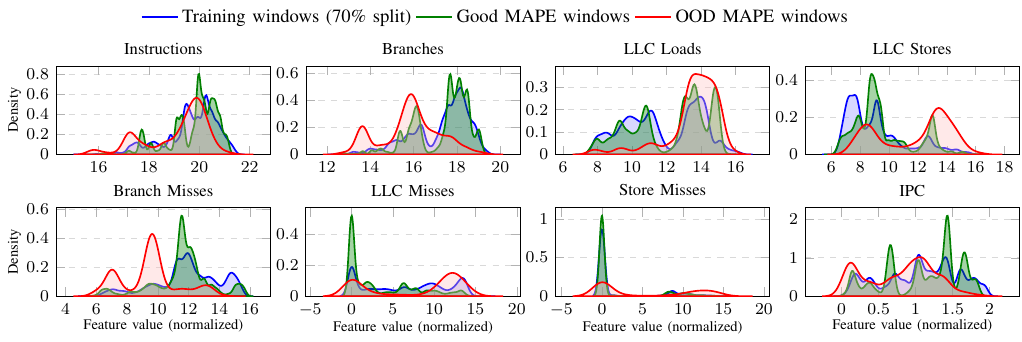}};
    \end{tikzpicture}
    \caption{
    KDE of eight raw PMC features across three window subsets: the 70\% training split~(blue), the test windows with a good MAPE below the 99th percentile~($\leq2.2$\%, green), and the OOD test windows in the 99th percentile tail~($>2.2$\%, red). 
    }
    \label{fig:kde_features_grid}
\end{figure*}

In this first experiment, the goal is to investigate the catastrophic behavior of the model on the \ac{OOD} inputs.
With one of the 50 model variants that was trained with a random 70\%/30\% split for training and testing, we use its 2.2\% 99th-percentile \ac{MAPE} as a classification threshold to classify test windows as either \textit{good} windows if the \ac{MAPE} is below the threshold, or as \ac{OOD} otherwise.
\Cref{fig:kde_features_grid} illustrates the \ac{KDE} of all eight \acp{PMC} used as model inputs, plotted separately for the 70\% training set~(blue), the \textit{good} test windows~(green), and the \ac{OOD} test windows~(red).
Two key findings emerge from this analysis.

\textbf{Finding 1: Model failure is not random.}
Across all eight features, the red distribution consistently shows a different density pattern compared to the blue training distribution. If failures are attributable to noise, the two distributions would overlap; this systematic shift shows that the model fails on a structurally-distinct subset of inputs, i.e., inputs with different underlying characteristics, and that silent failure on \ac{OOD} inputs is an inherent, systematic phenomenon in \ac{ML}-based resource management.

\textbf{Finding 2: No single feature detects failure.}
Despite this systematic displacement, the red and blue distributions overlap significantly in each individual feature.
Because of this overlap, no single \ac{PMC} threshold reliably separates the test windows in the 99th percentile \ac{OOD} windows from the \textit{good} ones.
Therefore, isolated feature monitoring is inadequate as a detection strategy.
Instead, any confidence estimator shall operate over a \textit{joint} representation that captures the multi-dimensional structure of the \ac{OOD} region.

\subsection{Embedding Space Analysis}
\label{sec:analysis:embeddings}

\begin{figure*}
    
    \centering
    \begin{tikzpicture}[x=1cm,y=1cm,font=\footnotesize,scale=0.9]

      \node[inner sep=0pt] (img) {\includegraphics[width=\linewidth]{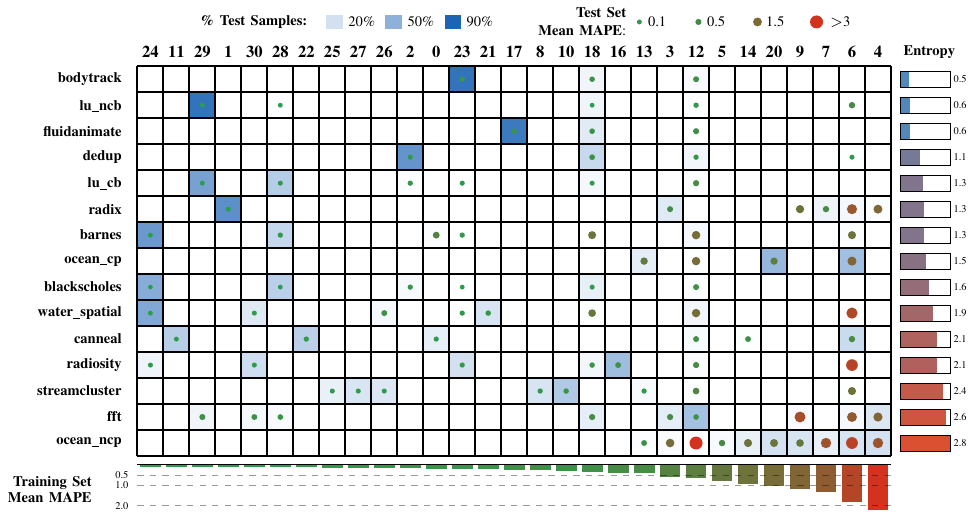}};
    \end{tikzpicture}
    \caption{
    Application behavioral fingerprint map.
    Each cell $(i,k)$ shows the fraction of application $i$'s test samples assigned to GMM cluster $k$ (blue intensity, row-normalized) and the mean MAPE achieved in that cell (dot color: green $\to$ red; dot size proportional to MAPE). }
    
    \label{fig:full-gmm-analysis}
\end{figure*}

To construct the \textit{joint} representations highlighted in Finding~2, embedding learning has emerged as a key enabling technique, originally developed and widely applied in the field of natural language processing.
These learned embeddings $\mathbf{e} = \mathrm{Enc}(\mathbf{x}) \in \mathbb{R}^h$ map each input to a latent space in which semantically-similar inputs are represented by nearby vectors~\cite{mikolov2013word2vec, devlin2019bert}. 
Applied to the \ac{IPS} forecasting usecase, instead of analyzing the individual features, embedding learning would yield a \textit{joint} dense representation of each input window, thereby capturing the phase-level characteristics of applications that incorporate all \acp{PMC} simultaneously.
With this representation, it is possible to analyze the model behavior on different windows based on their similarities, i.e., similarity between embedding vectors.
The underlying idea is that if the model can predict the \ac{IPS} of a window with a certain level of accuracy, it should be able to achieve similar accuracy when presented with \textit{similar} input windows.
In the following subsections, we analyze the effectiveness of embedding learning in capturing such similarities across applications and hardware configurations.

\subsubsection{Autoencoder for embeddings extraction}

We train an autoencoder on the complete dataset of periodic phase-level traces of all applications across the two core types and the 5 \ac{VF} levels.
The autoencoder uses a stacked LSTM-based architecture, comprising layers with 128 and 64 units (the embedding dimensions) using ReLU activations.
During training, the autoencoder learns a compressed representation of the data in an unsupervised manner and maps each input window to an $h$-dimensional latent vector.
Subsequently, the decoder uses this latent representation to reconstruct the input.
The goal during the autoencoder training is to minimize the reconstruction error between the input and the decoded output.
For this model, the achieved reconstruction error is 0.01667.

\subsubsection{t-SNE Visualization}
 
\begin{figure}
    
    \centering
    \begin{tikzpicture}
      \node[inner sep=0pt] (img) {\includegraphics[width=\linewidth]{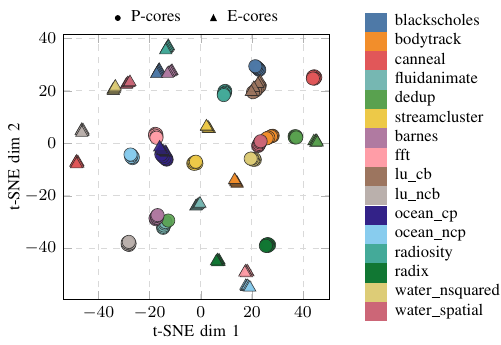}};
    \end{tikzpicture}

    \caption{
    t-SNE visualization of phase-level embeddings.
    Filled circle markers denote P-core executions; triangle markers
    denote E-core executions.
    Each point is the mean embedding of all phases of one
    application under one configuration.
    }
    \label{fig:tsne}
\end{figure}
We apply t-SNE~\cite{vandermaaten2008tsne} to reduce the learned embedding vectors to 2D for visualization and show the results in \cref{fig:tsne}.
Each point represents the mean embedding of all windows of one application under a given configuration, i.e., core type and \ac{VF} level, serving as a semantically-grounded anchor for the application's execution behavior under those conditions.

\textbf{Finding 3: Absorption of \ac{VF} Levels and Structural Separation by Core Type.}
For every application, the P-core and E-core embeddings form two clearly separate sub-clusters, reflecting the microarchitectural differences between the two core types.
This provides empirical evidence that the learned representations encode hardware context as a key dimension, independent of application identity. 
Moreover, for each pair of application and core type, the embeddings corresponding to the five \ac{VF} levels form a tight and compact cluster, indicating that the encoder captures a similar behavioral signature for each combination of application and core type that is preserved between operating frequencies, rather than modeling each level \ac{VF} as a distinct execution pattern.

\subsection{Cluster-Level Model Performance Analysis using GMM}
\label{sec:analysis:clusters}

Finding~3 confirms that the embedding space captures meaningful structure at both the application and hardware levels.
We aim to demonstrate that \ac{GMM} clustering of the learned embeddings can effectively approximate the target model behavior. Specifically, if the target model is exposed to a new phase assigned to cluster $k$ at runtime, the prediction error should exhibit statistical properties consistent with the historical error profile of that cluster.

To explore this direction, we fit a \ac{GMM} with $N=32$ components on the embeddings of all training phases and compute for each cluster $k$ the 90th-percentile \ac{MAPE} over the train set, serving as the per-cluster confidence estimate $\epsilon_k$.

\subsubsection{Behavioral Fingerprint Map}
\label{sec:behavioral-map}

\Cref{fig:full-gmm-analysis} visualizes cluster assignments and per-cluster model error across benchmark applications, with rows sorted by increasing behavioral entropy and columns by increasing cluster difficulty.
The figure highlights multiple patterns of predictability and complexity.
\textbf{First}, monobehavioral applications like \textit{bodytrack}, \textit{lu\_ncb}, and \textit{fluidanimate} concentrate over 86\% of their test windows in a single low-difficulty cluster, resulting in entropies below 0.6 and weighted-mean \ac{MAPE} scores under 0.3\%.
\textbf{Second}, moderately structured applications such as \textit{dedup}, \textit{lu\_cb}, \textit{radix}, \textit{barnes}, and \textit{blackscholes} have input windows distributed across two to five different clusters, but maintain dominant low-difficulty assignments of 60–73\%, yielding similarly low errors.
Yet, \textit{ocean\_cp} illustrates that moderate entropy does not always guarantee predictability, as its activity is confined to higher-difficulty clusters, producing a weighted-mean \ac{MAPE} of 1.22\%. 
\textbf{Third}, diffuse applications including \textit{canneal}, \textit{radiosity}, \textit{streamcluster}, \textit{fft}, and \textit{ocean\_ncp} spread windows broadly with no cluster exceeding 40\% occupancy, resulting in higher weighted-mean \ac{MAPE}, particularly in \textit{ocean\_ncp} (1.65\%), which cycles through distinct computational phases.
\textbf{Finally}, cluster 12 emerges as a universal way-point.
It is touched by all applications, occupies a mid-difficulty position and demonstrates that model performance within a cluster does not depend on the full application profile, but rather on the phase-level inputs of its execution. 

\textbf{Finding~4: Behavioral Entropy as a Strong Proxy for Model Performance Predictability}.
Low-entropy applications concentrate in \textit{easy} clusters and incur low \ac{MAPE}, i.e., good model performance, while high-entropy applications spread across the difficulty axis and incur higher prediction error, potentially indicating \ac{OOD}.
However, this entropy is independent of cluster assignment.
As demonstrated in the analysis of \textit{ocean\_ncp}, even applications that have a compact behavioral footprint can be hard to predict when all of their training input data falls in inherently difficult clusters.

This analysis demonstrates that \textit{cluster membership is a reliable predictor of model error, independently of application identity, confirming that \ac{GMM} clustering of the learned embeddings adds quantifiable information beyond any static global threshold, insights that can be leveraged to build a framework for dynamic per-input estimation of model prediction error at runtime.}
\section{\ourtechplain (\ourtech): Embedding- and GMM-based Confidence Estimation}
\label{sec:framework}

We propose \ourtechplain~(\ourtech), a post-hoc XAI framework that extends pre-trained \ac{ML} models ($f_\theta$) with a runtime confidence estimate $\mathcal{C}(\mathbf{x})$, without modifying their parameters or prediction logic.
\ourtech operates in two phases.
At \emph{design-time}, a \ac{GMM} is fitted to the embeddings learned by an encoder from the training data $\mathcal{D}_{\mathrm{train}}$, and a per-cluster statistical error \ac{MAPE} lookup table is constructed.
At \emph{runtime}, each incoming input is encoded, mapped to its most probable cluster, and the associated confidence estimate is retrieved before the prediction is issued.
\Cref{fig:overview} illustrates the full framework.

\subsection{Design-Time Procedure}
\label{sec:framework:design}

\textbf{The Encoder.}
Motivated by the findings in \cref{sec:analysis:failure} and \cref{sec:analysis:embeddings}, we adopt embedding learning to provide a rich joint representation of \acp{PMC}.
Let $\mathrm{Enc} : \mathcal{X} \rightarrow \mathbb{R}^{h}$ denote an encoder that maps any input $\mathbf{x} \in \mathcal{X}$ to a $h$-dimensional embedding vector $\mathbf{e} = \mathrm{Enc}(\mathbf{x})$.
For models whose internal representation is directly accessible, e.g., the output of an intermediate layer, $\mathrm{Enc}$ reuses that representation at no additional training cost.
For models that do not expose a suitable internal embedding, e.g., gradient-boosted trees, a lightweight neural network is trained,e.g., using a teacher–student framework, to approximate the output of the target model in $\mathcal{D}_{\mathrm{train}}$ and subsequently use its encoder to derive embeddings that effectively capture the underlying structural properties of the input space.

\textbf{GMM fitting.}
The encoder is applied to all inputs in $\mathcal{D}_{\mathrm{train}}$, yielding a set of embedding vectors $\mathcal{E}_{\mathrm{train}} = \{\mathrm{Enc}(\mathbf{x}) : \mathbf{x} \in \mathcal{D}_{\mathrm{train}}\}$.
A \ac{GMM} with $N$ components $\mathcal{G} = \{\pi_k, \bm{\mu}_k, \bm{\Sigma}_k\}_{k=1}^{N}$ is then fitted on $\mathcal{E}_{\mathrm{train}}$, partitioning the embedding space into $N$ clusters.
Here, \(\pi_k\) represents the mixing coefficient for the \(k\)-th Gaussian component, which is the prior probability of the \(k\)-th component in the mixture, \(\bm{\mu}_k\) is the mean vector of the \(k\)-th Gaussian component, and \(\bm{\Sigma}_k\) is the covariance matrix of the \(k\)-th Gaussian component, which characterizes the shape and orientation of the component's distribution.
Each cluster represents a subset of embeddings that are statistically similar under a shared Gaussian distribution, as demonstrated earlier in \cref{sec:analysis:clusters}

\begin{figure*}
    
    \centering
    \begin{tikzpicture}[x=1cm,y=1cm,font=\footnotesize,scale=0.9]

      \node[inner sep=0pt] (img) {\includegraphics[width=\linewidth]{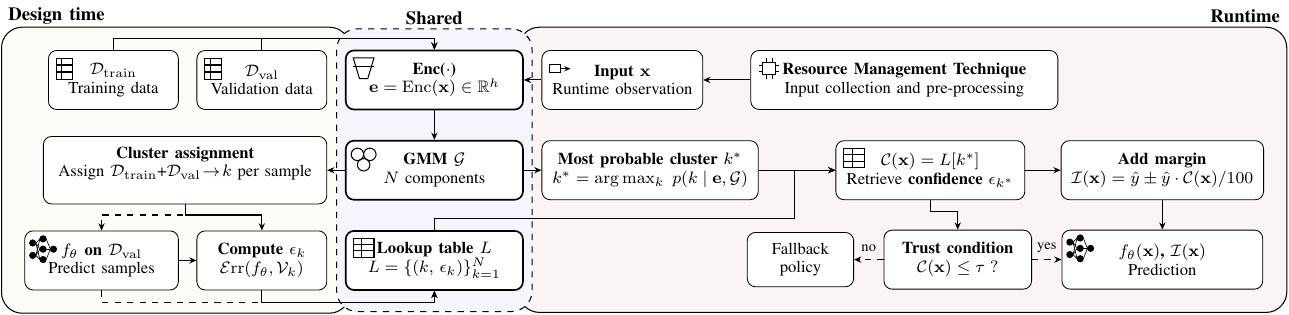}};
    \end{tikzpicture}
    \caption{Overview of \ourtechplain. At design time (left), an
      encoder $\mathrm{Enc}(\cdot)$ and a Gaussian Mixture Model
      $\mathcal{G}$ are fitted on the training embeddings; a per-cluster error lookup table $L$ is constructed from the validation set.
      At runtime (right), the incoming input $\mathbf{x}$ is encoded,
      mapped to its most probable cluster $k^*$, and the associated confidence
      estimate $\mathcal{C}(\mathbf{x})=L[k^*]$ is retrieved. The confidence is used
      in two ways: it is added as a margin to the prediction, yielding the interval
      $\mathcal{I}(\mathbf{x})$, and it is checked by the trust condition. The
      resource management technique uses the prediction with its interval
      when $\mathcal{C}(\mathbf{x})\leq\tau$,
      or falls back to a conservative policy otherwise. The encoder,
      GMM and lookup table are shared across both phases (middle purple-shaded box).}
\label{fig:overview}

\end{figure*}

\textbf{Lookup table construction.}
The model $f_\theta$ is queried for all inputs in $\mathcal{D}_{\mathrm{train}}$ and the prediction error is measured for each input.
For each cluster $k$, the per-cluster confidence estimate $\epsilon_k$ is derived by computing a representative statistic of the prediction error over all training samples assigned to that cluster, e.g., the 90th-percentile \ac{MAPE} for regression tasks or the misclassification rate for classification tasks.:
\begin{equation}
    \epsilon_k = \mathcal{E}\text{rr}(f_\theta,\, \mathcal{V}_k),
    \quad
    \mathcal{V}_{k^*} = \{(\mathbf{x}, y) \in \mathcal{D}_{\mathrm{train}} ,
    k^* = k\}
    \label{eq:epsilon}
\end{equation}
If no training input is assigned to cluster $k$, we set $\epsilon_k = \infty$, indicating that there is no historical error estimate available for that region of the input space.
The resulting lookup table $L = \{(k, \epsilon_k)\}_{k=1}^{N}$ is stored for use at runtime.
The trained encoder ($\mathrm{Enc}$), the fitted GMM ($\mathcal{G}$), and the resulting lookup table ($L$) are utilized at runtime to estimate error bounds for new inputs, as depicted in the shared partition of \Cref{fig:overview}.

\subsection{Runtime Procedure}
\label{sec:framework:runtime}

At each prediction step, \ourtech intercepts the incoming input before it is processed by $f_\theta$ and produces a confidence estimate by first identifying the most probable cluster for the embedding of the input, 
$k^* = \arg\max_{k}\ p(k \mid \mathbf{e}, \mathcal{G})$, and then retrieving the associated $\epsilon_k$ from the lookup table $L$.
Afterward, a trust condition, predefined by the resource management technique, is evaluated \emph{before} querying $f_\theta$: if the trust condition is violated, a fallback policy is invoked immediately without incurring the cost of a catastrophic model inference.

\textbf{Confidence interval.}
When the trust condition is satisfied, \ourtech returns both the prediction $\hat{y}$ and a symmetric confidence interval:
\begin{equation}
    \mathcal{I}(\mathbf{x}) =
    \Bigl[
        \hat{y} \cdot \Bigl(1 - \tfrac{\mathcal{C}(\mathbf{x})}{100}\Bigr),\;
        \hat{y} \cdot \Bigl(1 + \tfrac{\mathcal{C}(\mathbf{x})}{100}\Bigr)
    \Bigr],
    \label{eq:interval}
\end{equation}
providing the resource management technique with an explanation that can be used to reason about the worst-case deviation of the prediction from the ground truth.

\section{Experimental Evaluation}
\label{sec:results}

To demonstrate the effectiveness of our method, we conduct an extensive set of experiments on the same Intel Alder Lake 24-core Linux-based machine, using the same target state-of-the-art \ac{IPS} forecasting model, detailed in sections \ref{sec:intro:motivational} and \ref{sec:analysis}.
Given the architecture of the \ac{LSTM}-based model, our \ourtech framework uses the first two layers as the encoder component, from which embeddings are directly extracted to fit the \ac{GMM}.
The hyperparameters of the \ac{GMM} are optimized via the grid search procedure. 
The lookup table is constructed using the 90th-percentile of the \ac{MAPE} values of training points in each cluster.

\subsection{Evaluation Strategy}

Experiments are designed to evaluate the robustness of our \ourtech framework in two dimensions of increasing complexity. The first dimension is software-level generalization, where models are trained and evaluated using the \ac{LOO} strategy either within the same benchmark suite (PARSEC and SPLASH-2) or on an unseen benchmark suite (SPEC CPU2006\cite{henning2006spec}). The second dimension is hardware-level generalization, where models are trained on either one core type (P core only) or on both core types together (P and E) across the available \ac{VF} levels. The \ac{LOO} strategy, described in \cref{sec:intro:motivational}, is used in all experiments. In each fold, the \ac{GMM} hyperparameters are selected using the Bayesian Information Criterion (BIC), computed on the training applications only, through a grid search over the number of clusters $N \in \{16, 32, 64, 128\}$ and the covariance regularization $\Sigma \in \{10^{-4}, 10^{-6}, 10^{-9}\}$. The lowest BIC is obtained with $N = 128$ and $\Sigma = 10^{-9}$ in every fold, which indicates that a fine partition of the latent space is needed to capture the diversity of execution phases. The performance of our technique is measured by the containment rate $\rho$, calculated as the proportion of predictions that fall within the predicted confidence interval $\epsilon_k$, and is compared against baselines that employ global model performance indicators.

\subsection{Comparison Baselines}
\label{sec:evaluation-baselines}

    


\begin{figure*}
  \centering
  \includegraphics[width=\linewidth]{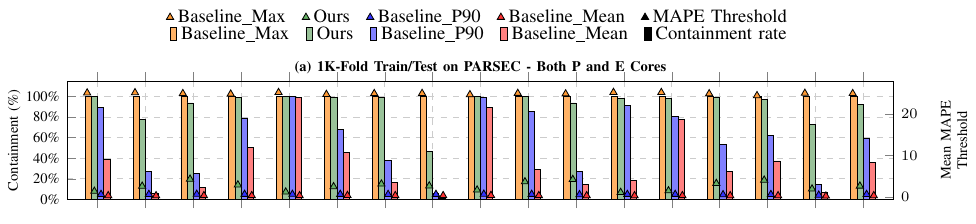}\\[0.5em]
  \includegraphics[width=\linewidth]{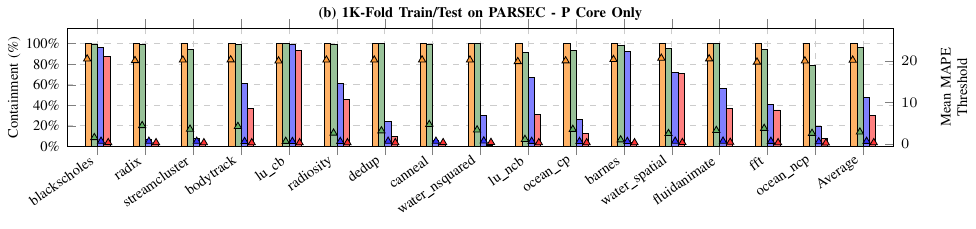}\\[0.5em]
  \includegraphics[width=\linewidth]{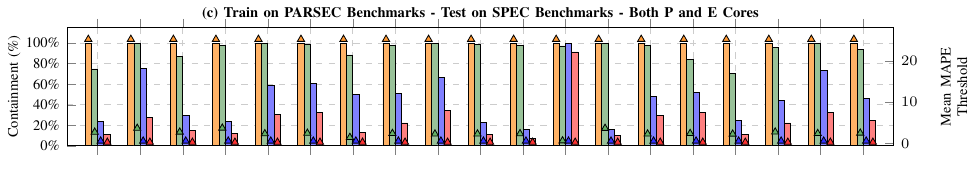}\\[0.5em]
  \includegraphics[width=\linewidth]{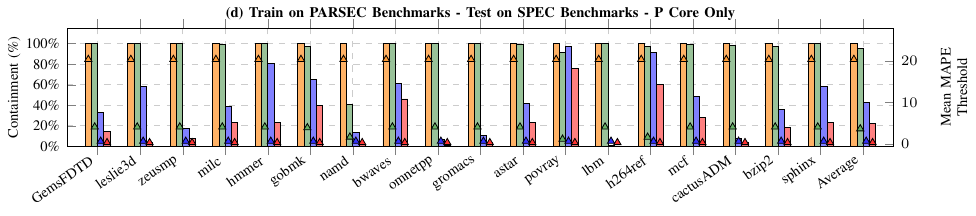}

  \caption{
      Performance comparison. Bars (left axis) show normalized performance for
      Max, Mean, P90, and our \ourtech. Triangle markers (right axis) show the corresponding
      absolute MAPE threshold values.
      Across the four experiments, our \ourtech consistently achieves an average performance of \textasciitilde94\%, on par with the Max baseline, i.e., the most pessimistic, while providing error margins that are \textasciitilde17\% tighter.
      This highlights the effectiveness of our model’s adaptive and input-specific confidence estimations.
  }
  \label{fig:all_results}
\end{figure*}

Our \ourtech framework is evaluated against three baseline approaches for handling model performance when faced with \ac{OOD} inputs.
Each baseline relies on one global error metric, specifically the maximum, mean, or 90th-percentile \ac{MAPE} score across the train dataset, applied uniformly across all input windows, denoted as \texttt{baseline\_max}, \texttt{baseline\_mean}, and \texttt{baseline\_P90}, respectively.

In contrast, our \ourtech method is input-specific: input windows are encoded, clustered, and assigned 90th‑percentile \ac{MAPE} bounds.
To demonstrate the robustness of our method across diverse benchmarks and hardware configurations, we conduct four distinct experiments.

\textbf{In the first experiment,} we employ the dataset comprising P core measurements exclusively.
This preliminary experiment is designed to evaluate the ability of our method to generalize to different applications within the same benchmark suite but on the same core type.
As illustrated in \Cref{fig:all_results}.b, our approach yields consistently solid results, attaining an average containment rate $\rho$ of 96\% across all dataset variants, which is comparable to the \texttt{baseline\_max} $\rho$ of 99\%.
However, the mean error margin considered by \ourtech in these experiments is 2.87\%, approximately an order of magnitude smaller than the \texttt{baseline\_max} error margin of 20.28\%.
In contrast, while the \texttt{baseline\_P90} strategy also reduces the error margin, it leads to a substantial degradation in predictive performance, nearly halving the $\rho$ from 96\% to 47\%.

\textbf{In the second experiment,} we increase the test scenario complexity by using a dataset combining both P and E core measurements, testing generalization across heterogeneous hardware.
As shown in \Cref{fig:all_results}.a, the model achieves 92\% average $\rho$ with a narrow error margin, thereby providing the best balance between accuracy and error margin compared to the \texttt{baseline\_max} and \texttt{baseline\_P90} approaches.
However, certain applications, e.g., \textit{canneal}, fall below 50\% accuracy, likely due to atypical behavior on E cores compared to P cores.
The two baselines \texttt{Baseline\_P90} and \texttt{baseline\_mean} struggled similarly, suggesting that the challenge stems from the application characteristics rather than a specific model limitation.
This confirms the highlighted behavior in \cref{sec:behavioral-map}, where some diffuse applications were identified, for which the \ac{GMM} assigned input windows to multiple clusters, but to none with more than 40\% occupancy.

\textbf{In the third experiment}, we further raise the complexity of the learning problem to evaluate \ourtech’s ability to generalize to unseen benchmarks, by training the model on PARSEC and SPLASH-2 benchmarks, but testing on SPEC CPU2006, considering only P core measurements. 
As shown in \Cref{fig:all_results}.d, our \ourtech maintains strong $\rho$, with 95\% average accuracy and a low adaptive mean error margin of 3.65\%.
This confirms our method’s robustness across diverse applications and its effective generalization to new benchmarks.
The application \textit{namd} again shows behavior similar to that of \textit{canneal} in the previous experiment.

\textbf{In the fourth experiment,} we build on top of the same evaluation strategy as in the previous experiment by using both P and E core measurements.
This experiment is the most challenging of the four, as it assesses the generalization capabilities of our \ourtech across both heterogeneous benchmarks \textit{and} heterogeneous hardware configurations. 
The model was trained on the PARSEC and SPLASH-2 datasets with aggregated P and E core data and then tested on each SPEC application individually.
As shown in \Cref{fig:all_results}.c, our \ourtech consistently achieves strong predictive performance, with an average $\rho$ of 93\% and an average mean \ac{MAPE} error bound of 2.49\%.
These results demonstrate that our method is robust across diverse applications and hardware configurations and generalizes even to heterogeneous scenarios.

\textbf{Overall}, these results demonstrate that our proposed approach successfully balances containment rate and error bounds.
This strength stems from our framework's ability to adapt to each individual input, leveraging performance on similar training windows to inform its error bound estimates.
As a result, the method generates measures tailored to each input, providing a reliability signal to the resource management technique to make reliable and well-informed decisions.

\subsection{Illustrative Example}
\label{sec:illustrative_example}

\begin{figure}
    
    \centering
    \begin{tikzpicture}
      \node[inner sep=0pt] (img) {\includegraphics[width=\linewidth]{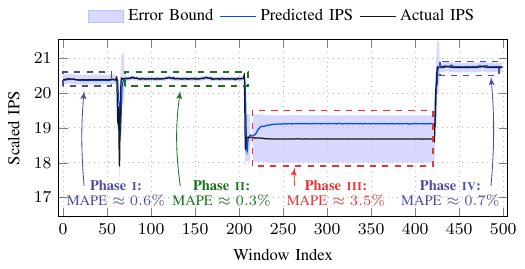}};
    \end{tikzpicture}

    \caption{
    t-SNE visualization of phase-level embeddings.
    Filled circle markers denote P-core executions; triangle markers
    denote E-core executions.
    Each point is the mean embedding of all phases of one
    application under one configuration.
    }
    \label{fig:illustrative}
\end{figure}

To gain insights into how our \ourtech adapts to various execution phases of applications at runtime, \cref{fig:illustrative} shows a detailed execution trace of \textit{water\_spatial} along with the actual \ac{IPS}, the predicted \ac{IPS} of the original model, as well as the predicted error bounds predicted by our \ourtech framework. 
As illustrated in the figure, the trace of \textit{water\_spatial}'s number of instructions is divided into distinct phases, where the original model's performance clearly fluctuates across phases.

Our method adapts successfully to these fluctuations.
Indeed, \ourtech predicts a small error margins in Phases \MakeUppercase{\romannumeral 1}, \MakeUppercase{\romannumeral 2}, and \MakeUppercase{\romannumeral 4}, which point to a high degree of model confidence.
This is also evident in the predicted \ac{IPS} values, which align closely with the actual \ac{IPS} observed in these phases.
In contrast, a substantially increased error margin is predicted by our method in Phase \MakeUppercase{\romannumeral 3}, where the original model's predicted \ac{IPS} values are approximately 10 times worse than in Phase \MakeUppercase{\romannumeral 2}.
Nonetheless, our method still adequately reflects this higher error by correspondingly broadening the error margin.

It is also important to note that our method allocates larger error bounds to the transition phases, which usually exhibit increased variability and a greater chance of prediction inaccuracies.
This behavior illustrates our \ourtech’s ability to adjust its error bound estimates in accordance with changes in phase complexity at runtime.

\section{Related Work}
\label{sec:related}

Techniques for open systems~\cite{open_system} assume that previously unknown applications may run at arbitrary, unpredictable times. Prior work in this area shows differing awareness of the need to evaluate models’ runtime generalization to unseen or \ac{OOD} scenarios.

The first category of approaches supports its generalization claims by leveraging the neural network's inherent ability to learn from limited training data and extrapolate to novel situations. 
These models are often evaluated by randomly splitting the available data, typically into a 70/15/15 train/validation/test split \cite{Edun2019related}, \cite{Gupta_rw_70-30_2020}, or a 70/30 train–test split \cite{Erika2022realted}, \cite{Kundan_rw_70-30_2020}.
These studies fail to demonstrate generalization due to insufficient testing on unseen workloads.

A second class of ML-based techniques examines model generalization. For instance, the evaluation protocol employed by the authors of \cite{Ribeiro_rw_cv_2022} relies on 10-fold cross-validation. 
This procedure doesn't guarantee that test folds only contain unseen applications, limiting cross-application generalization.
To address this limitation, \cite{contribution_smartboost_dac_2021}, \cite{Narang_rw_cv_2023}, and \cite{contribution_efficast} evaluated model performance on applications excluded from the training dataset. 
Despite improvements, these approaches focus on application-level generalization and overlook phase-level variability, risking performance degradation.

To mitigate prediction confidence, the authors of \cite{contribution_tcrm}, \cite{Chen_rw_em_2023} propose using global error bounds based on statistics derived from the training data. 
Complementing these strategies, the work of \cite{Wachter_rw_em_2019} relies on a lookup table that captures discrepancies between predicted and observed values for each operating condition.
As demonstrated in \cref{sec:intro:motivational} and further corroborated in \cref{sec:results}, such global confidence mechanisms yield acceptable performance only when a highly pessimistic margin is applied, indicating that an input-specific technique is required.

Recent work explores embedding-based representations to improve interpretability and quantify model confidence in various AI applications.
In \cite{brunner2019semanticembeddings}, the authors propose a Semantic Autoencoder that projects network representations into an attribute space, using cosine distance to quantify confidence and selectively reject predictions with large semantic distances, enhancing model reliability.
Similarly, the authors of \cite{macdonald2025LLMembedding} extract dense embedding vectors from large language models' hidden representations, normalize input sequences, and train an auxiliary classifier to estimate confidence in the generated output.
The authors of \cite{wu_rw_emb_2025} fit a \ac{GMM} to the pairwise distances in the training-data embedding space, using it to identify \ac{OOD} samples and enhance the robustness and performance of GUI agents.
In addition, the authors of \cite{Moradi_rm_embd_2021} introduced confident itemset collections to discretize the feature space, extracting correlations to generate class-wise and instance-wise explanations that align with the model's behavior.
These studies show that embedding-based representations offer both interpretability and calibrated confidence, enhancing the trustworthiness of AI predictions.

\section{Conclusion}

This work addresses the challenge of reliable confidence estimation in \ac{ML}-based resource management under distributional shifts, where methods in the literature that use a global error margin failed to capture input-dependent variations at runtime. To this end, we proposed \ourtech, a post-hoc XAI framework that learns an embedding space, clusters it with a \ac{GMM}, and utilizes statistical error estimates for confidence signals that are input specific rather than global. Evaluated on a state-of-the-art \ac{LSTM}-based \ac{IPS} forecasting model, \ourtech achieves 94\% containment rate across 34 benchmark applications and hardware configurations, providing 17\% tighter confidence bounds than global baselines with negligible runtime overhead, showcasing the effectiveness of input-dependent confidence estimation in enhancing the reliability of ML-based resource management.

\bibliographystyle{ACM-Reference-Format}
\bibliography{references}

\end{document}